\documentclass[conference]{IEEEtran}
\IEEEoverridecommandlockouts

\usepackage{graphicx}
\usepackage{color}
\usepackage{stfloats}
\usepackage{booktabs}
\usepackage{cite}
\usepackage{amsmath,amssymb,amsfonts}
\usepackage{algorithmic}
\usepackage{graphicx}
\usepackage{textcomp}
\usepackage{xcolor}

\newcommand{\todo}[1]{{#1}}

\def\BibTeX{{\rm B\kern-.05em{\sc i\kern-.025em b}\kern-.08em
    T\kern-.1667em\lower.7ex\hbox{E}\kern-.125emX}}

\begin{document}

\title{Semi-Supervised Self-Taught Deep Learning for Finger Bones Segmentation\\}
 \author{\textit{Ziyuan Zhao\thanks{* The first two authors contribute equally in this work.}$^{*1,2}$, Xiaoman Zhang$^{*1,2}$, Cen Chen$^1$, Wei Li$^4$, Songyou Peng$^1$}\\
\textit{Jie Wang$^1$,  Xulei Yang$^3$, Le Zhang$^1$ and Zeng Zeng$^1$} \\ 
$^1$\textit{Institute for Infocomm Research, A*STAR, Singapore} \\ 
$^2$\textit{National University of Singapore, Singapore}  \\
$^3$ \textit{YITU Tech, Singapore} \\
$^4$ \textit{Huazhong University of Science and Technology, China} \\}
 \maketitle

\begin{abstract}

Segmentation stands at the forefront of many high-level vision tasks. In this study, we focus on segmenting finger bones within a newly introduced semi-supervised self-taught deep learning framework which consists of a student network and a stand-alone teacher module. The whole system is boosted in a life-long learning manner wherein each step the teacher module provides a refinement for the student network to learn with newly unlabeled data. Experimental results demonstrate the superiority of the proposed method over conventional supervised deep learning methods.
\end{abstract}

\section{INTRODUCTION} \label{introduction}
We study a fundamental problem of finger bones segmentation which
has broad applications in clinical practice such as bone age assessment~\cite{gertych2007bone}, \todo{anomaly detection~\cite{dominguez2008detection}} and so on. Compared with other segmentation \todo{tasks~\cite{liu2018deep}} in the computer vision community, finger bones segmentation is challenging due to to the variation of hand bones and scarceness of large-scale well-annotated dataset.



As other pixel-wise regression tasks in computer vision~\cite{liu2017richer}, image segmentation is an important area of research in medical image analysis where many attempts and success have been made so far. In the domain of medical image analysis, on the one hand, thresholding based segmentation methods were widely applied. These methods assume that the background and foreground of an image have distinct intensity range. Pietka~\emph{et al.}~\cite{pietka2001computer} used the Sobel gradient to segment bones with soft tissue region. Similar techniques like Derivative of Gaussian (DoG) and dynamic thresholding have also been extensively studied for hand bone images~\cite{pietka1993feature,sharif1994bone,giordano2007epiphysis}. However, the performance of these methods is not satisfying on the radiographs without bi-model histogram. On the other hand, clustering-based approaches have also been applied on segmentation~\cite{zhang2007automatic,tristan2008radius,giordano2010automatic} and they are shown to perform well on images with a low-intensity range. However, it is difficult to generate relatively stable masks for different bones due to significant variations in the maturity of bone and image quality~\cite{van2018automated}.

Collecting a large scale dataset containing all challenging scenarios of fingers, as a common practice in the vision community, may partially alleviate the mentioned challenges. However,  manual labeling is costly, time-consuming, error-prone,  and requires massive human intervention for each new task. This is often impractical, especially for clinical practices due to some privacy concerns. This further motivates us to study the following question: \emph{Is it possible to improve the finger bone segmentation performance with the unlabeled dataset? }

In this paper, we propose a novel self-taught deep learning framework which consists of a student network and a stand-alone teacher module. The U-Net is firstly trained on a small subset of well-annotated training images and then boosted in a life-long learning manner. In each step, the teacher module provides a refinement for the student network to learn with newly unlabeled data. Experimental results show that the proposed method outperformed conventional supervised deep learning methods.


\section{RELATED WORK}

In recent years, the number of success stories of segmentation has seen an all-time rise\cite{shi2018bayesian,shi2018crowd}. The unifying idea behind all of the above is deep learning, the utilization of neural networks~\cite{long2015fully} with many hidden layers, for the purposes of learning complex feature representations from raw data, rather than relying on handcrafted feature extraction. They have shown consistent improvements over their non-deep counterparts across many tasks beyond segmentation\cite{zhang2018persemon,zhao2019Contrast}. Those approaches usually adopt an ``encoder-decoder" structure which could gradually decrease the resolution of the input with the depth of the network in the encoding stage, and then up-sampling and skip connections are applied to recover the resolution of the input in the decoding stage. So far, much of the recent research on segmentation using this kind of network has been made~\cite{badrinarayanan2015segnet,ronneberger2015u,lin2017refinenet, zeng2018multi}. In particular, U-Net \cite{ronneberger2015u} is the most widely used in biomedical image segmentation. It concatenates multi-scale feature maps in the encoding stage to upsample feature maps in the decoding stage. This design helps to generate richer feature hierarchies and achieve outstanding performance under the condition of very few annotated images. Although much progress has been made, the results from deep networks are not ideal from a practical point of view because the boundary information of the resulting segmentation maps may be lost for some complex background. To address this, the predicted results are usually further refined with an additional graphical model, for instance, a fully connected CRF, in which, pixels treated as nodes are pairwise connected with each other~\cite{krahenbuhl2011efficient}.



\section{Methodology} \label{methodology}
In this section, we explore the semi-supervised self-taught deep learning pipeline proposed for pseudo-labeling and finer segmentation. 

\subsection{Preprocessing and Augmentation}
\begin{figure}
    \centering
    \includegraphics[scale=0.35]{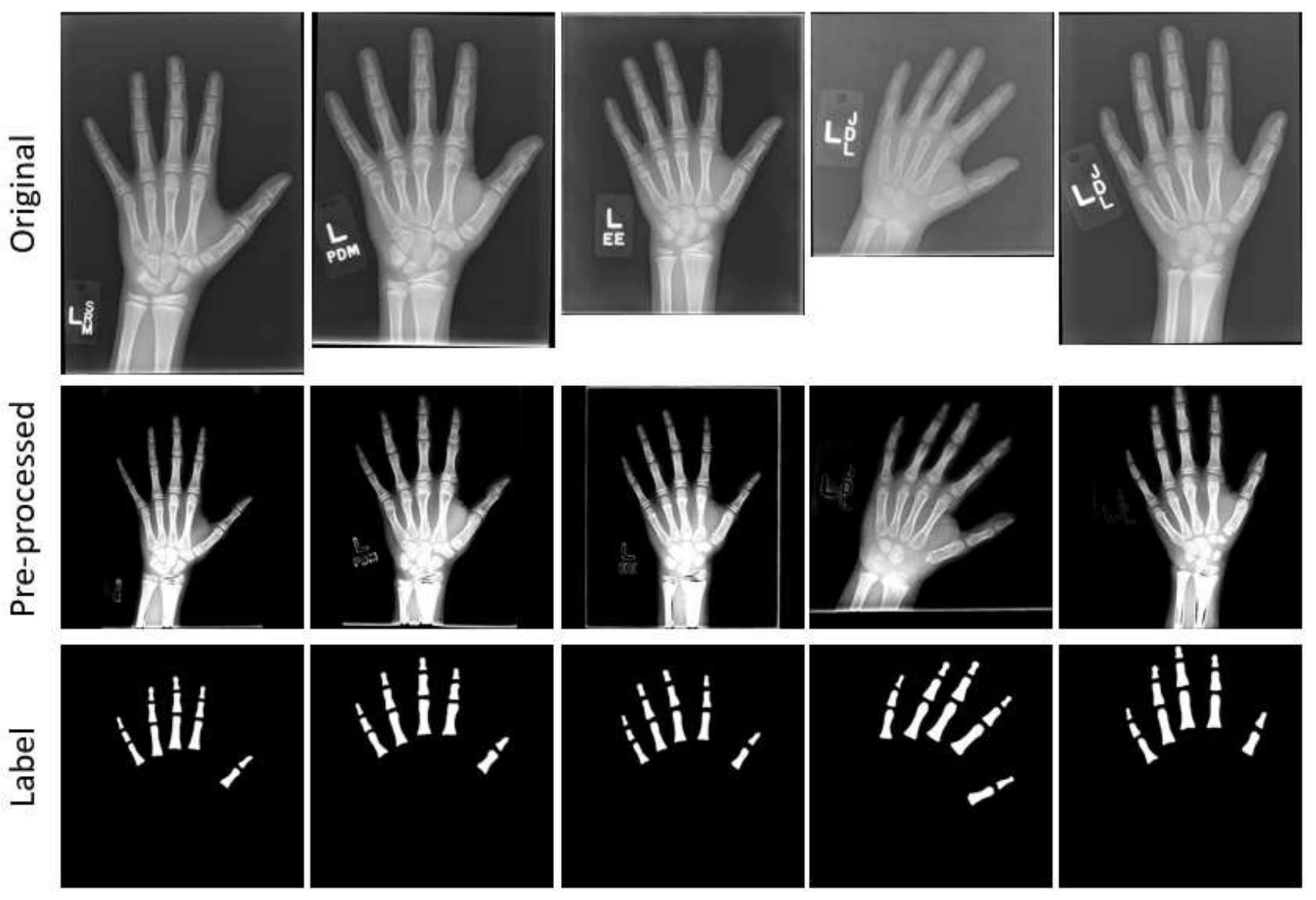}
    \caption{Examples of preprocessing:(first row) original images; (second row) preprocessed images; (third row) binary finger bone masks}
    \label{fig:preprocess}
\end{figure}

Intensity normalization was first applied by subtracting the mean and then dividing by the standard deviation to improve the brightness and augment the edges of bones. After that, radiographs were cropped and scaled to a fixed size with 600 x 600 pixels. Finally, the ground truth masks of these 209 radiographs \todo{selected from Section~\ref{dataset}} were annotated manually for the following experiments, in which, phalanges(Distal, Middle, Proximal) were annotated. Some examples are shown in Fig~\ref{fig:preprocess}.
\todo{To enhance the data, the images were randomly rotated by $\pm$20 degrees, shifted in width and height by 0.05, zoomed by 0.05, sheared by 0.05 and flipped horizontally.}

\subsection{Self-taught Learning Pipeline}
\begin{figure*}[htb]
    \centering
    \includegraphics[scale=0.68]{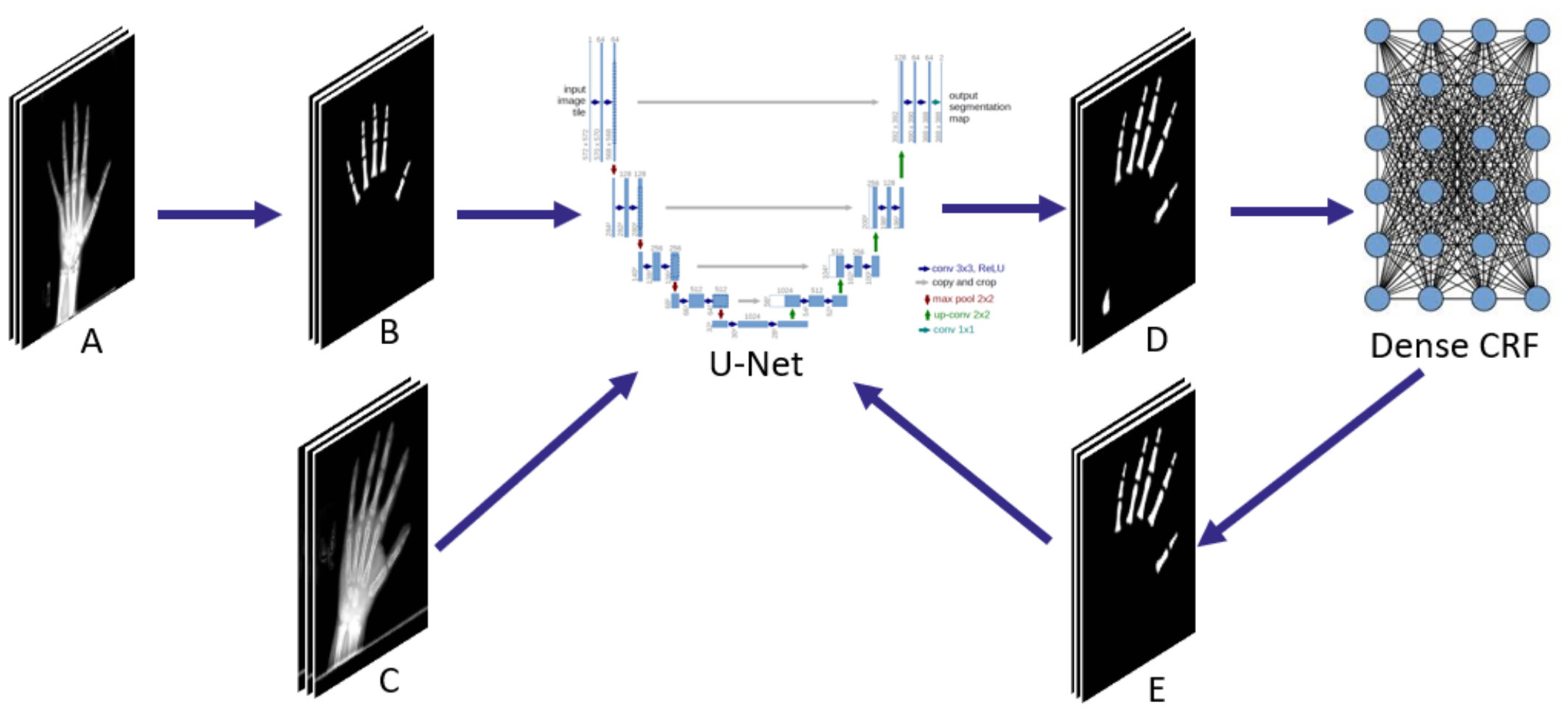}
    \caption{The iterative procedure of self-taught learning utilizing U-Net and Dense CRF for pseudo-labeling: (A) preprocessed input data; (B) masks manually labeled; (C) new data; (D) raw prediction; (E) refine prediction. The system boosts itself in a “Curriculum Learning” manner.  More specifically, in each iteration,  the  CRF  refines the raw predictions of U-Net and return the results from the easiest samples therein as pseudo-labels for future learning.}
    \label{fig:pipeline}
\end{figure*}

As shown in Fig~\ref{fig:pipeline}, the proposed system consists of a student module, which is realized by a Deep U-Net, and a teacher module embodied by a dense CRF.  First, the U-Net is trained firstly on a small set of the well-annotated training set and makes predictions on the unlabeled.  After that, a stand-alone dense CRF module is utilized to make refinements and provide pseudo-label for the U-Net. 

More specifically, every pixel $i$,  which is regarded as a node, has a label $x_{i}$ and an observation value $y_{i}$, and the relationship among pixels are regarded as edges. The labels behind pixels can be inferred by observations $y_{i}$, and the dense CRF $I$ is characterized by a Gibbs distribution,

\begin{equation}
P\left ( X=x|I \right ) = \frac{1}{Z(I)}exp(-E(x|I))
\end{equation}
where $E(x|I)$ is the Gibbs energy of a label $x$, which is formulated as follows,
\begin{equation}
E(x) = \sum \limits_{i}\Psi _{u}\left ( x_{i} \right )+\sum\limits_{i<j}\Psi _{p}(x_{i},x_{j})
\end{equation}
among which, the unary potential function $\Psi _{u}\left ( x_{i} \right )$ is donated by the output of U-Net, and the pairwise potentials in our model is given by
\begin{equation}
\Psi _{p}(x_{i},x_{j})=\mu(x_{i},x_{j})\sum\limits_{m=1}^{M}w^{(m)}k_{G}^{(m)}(f_{i},f_{j})
\end{equation}
where each $k_{G}^{(m)}$ is a Gaussian kernel $k_{m}(f_{i},f_{j})$, the vectors $f_{i}$ and $f_{j}$ are feature vectors for pixels $i$ and $j$, $w^{(m)}$ are linear combination weights, and $\mu$ is a label compatibility function.

Our system iterates in a ``curriculum Learning" manner in which easier samples are firstly chosen to improve the U-Net and difficult samples are gradually included. In each iteration, we choose $N$ easiest samples and use the results from dense CRF as the pseudo-label for U-Net.  More specifically, we calculate the Dice's Coefficient, as defined in Eq~\ref{dice} where $X$ and $Y$ are the cardinalities of the two sets, between the input and output of dense CRF. Dice's Coefficient serves as a proxy of the difficulty for each sample. A larger Dice's Coefficient indicates that U-Net performs relatively well because no significant refinements are given by dense CRF. In this way, more reliable information is first utilized in the system, which provides better audiences for the U-Net to learn.
\begin{equation}
  DSC = \frac{2|X\cap Y|}{|X|+|Y|}
  \label{dice}
\end{equation}

\section{EXPERIMENTS}\label{Experiments}
\subsection{Dataset and Evaluation Protocol} \label{dataset}

\begin{figure}[htb]
    \centering
    \includegraphics[scale=0.3]{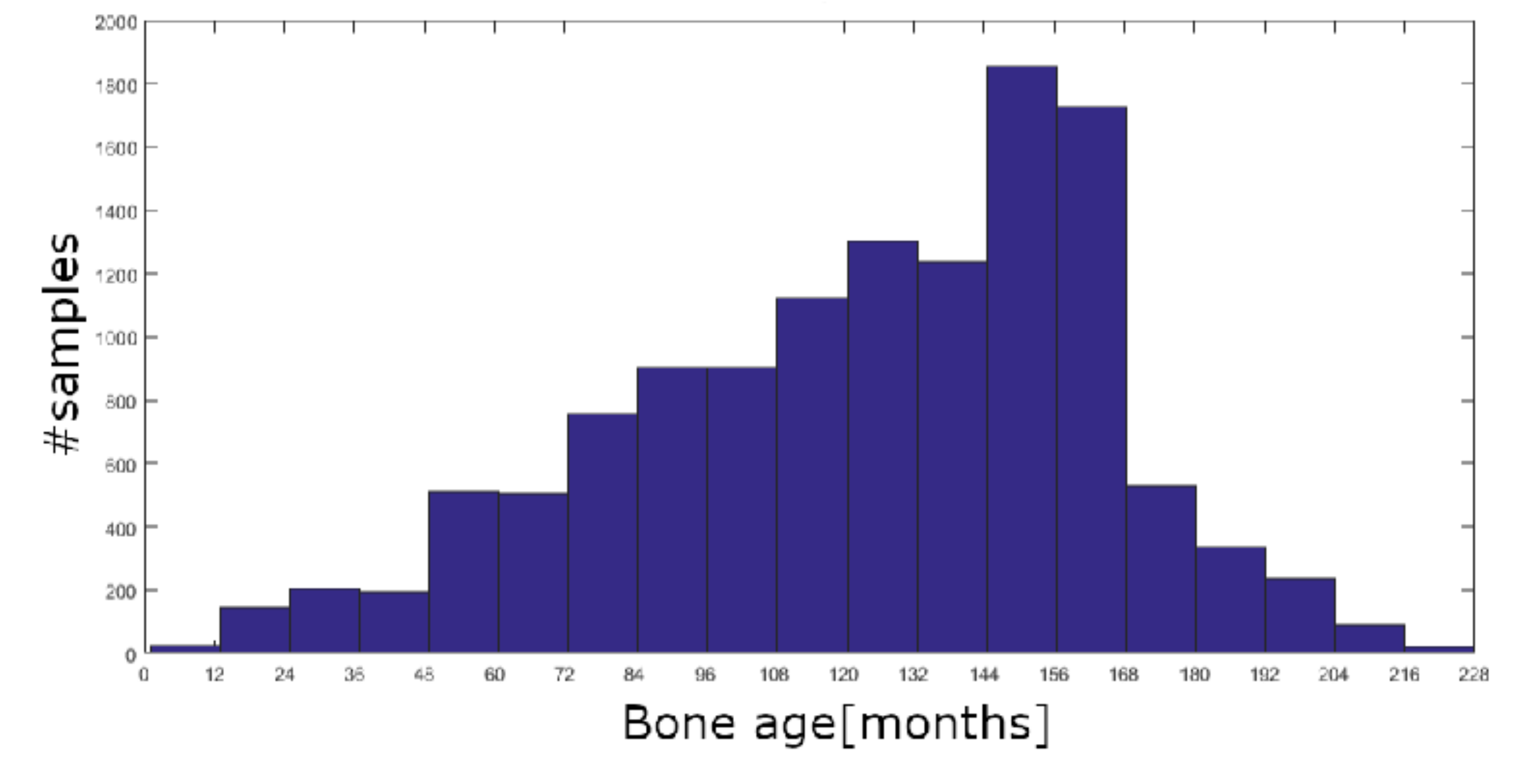}
    \caption{Distribution of estimated bone age.}
    \label{fig:agedist}
\end{figure}

\begin{figure}[htb]
    \centering
    \includegraphics[scale=0.4]{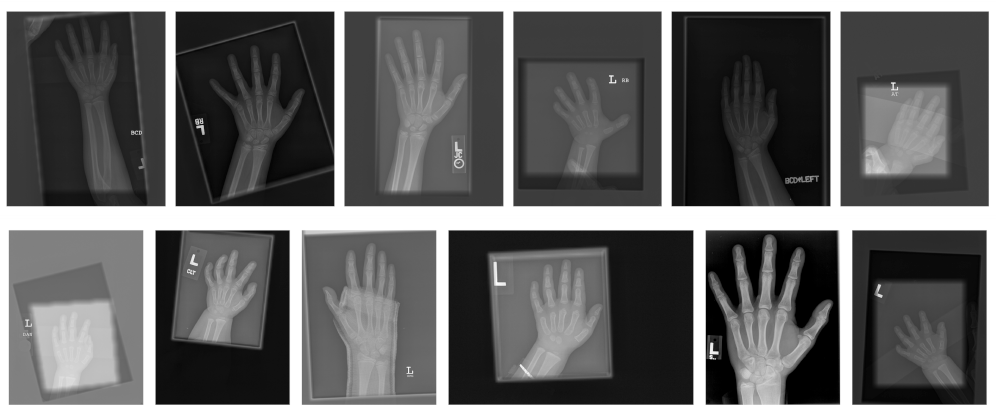}
    \caption{Example of radiographs in the dataset in different size, contrast,brightness.And some shows additional artifacts.}
    \label{fig:examples}
\end{figure}

The dataset used in this paper is from the 2017 Pediatric Bone Age Prediction Challenge~\cite{larson2017performance} organized by the Radiological Society of North America (RSNA). It consists of 12611 radiographs of the hand containing 6833 images of male and 5778 images of the female. The overall age distribution of this dataset is severely imbalanced, as shown in Fig~\ref{fig:agedist}. And Fig~\ref{fig:examples} shows some examples of radiographs in the dataset, and hand bones differ significantly from size, brightness, orientation and contrast across the samples. Besides, some artifacts are shown on the radiographs, such as watches and plaster casts.

Due to the size of the dataset and time consuming, small batch of the dataset is applied to our experiments to validate the effectiveness of the method proposed. To have a suitable generalization of the segmentation, the dataset was grouped in 19 year-based age groups (0-1, 1-2, 2-3, \ldots ,18-19), then 11 cases were selected from each year group to form the small subset of Training Set ($11\times19 = 209$) for experiments.Finally, the dataset is  split a into a train (139) /validation (20) /test (50) set randomly.



To validate the effectiveness of the pipeline we proposed, experiments were done in 209 images with ground truth masks, which is further randomly split into 4 subsets: training, validation, test and one 50-size subset prepared to enhance the training data. The U-Net has an input size of 512 $\times$ 512 pixels and is optimized with Dice's Coefficient. Hyper-parameters of dense CRF are tuned on the training set (detailed shown in Table~\ref{le_crfparam}). In this study, we set $N=10$ and compare our system with the three following baselines:
\begin{enumerate}
    \item U-Net-89: U-Net trained with 89 training samples.
    \item U-Net-139: U-Net trained with 139 training samples.
    \item \todo{U-Net-Only: U-Net trained with 89 training samples. After that, we randomly select 10 results of U-Net as the pseudo labels.}
\end{enumerate}



\begin{table}[htb]
\caption{Hyperparameters of Dense CRF}
\label{le_crfparam}
\begin{center}
\begin{tabular}{|c|c|c|c|c|}
\cline{1-5}
                                       & \multicolumn{1}{c|}{sdims} & \multicolumn{1}{l|}{schan} & \multicolumn{1}{l|}{compat} & \multicolumn{1}{l|}{step} \\ \hline
\multicolumn{1}{|c|}{PairwiseEnergy}   & 10                         & 10                         & 3                           & -                         \\ \hline \cline{1-1}
\multicolumn{1}{|c|}{PairwiseGaussian} & -                          & -                          & 3                           & -                         \\  \cline{1-5}
\multicolumn{1}{|c|}{inference}        & -                          & -                          & -                           & 50                        \\ \hline
\end{tabular}
\end{center}
\end{table}
\subsection{Results and Discussions} \label{results}

\begin{figure}[htb]
    \centering
    \includegraphics[scale=0.5]{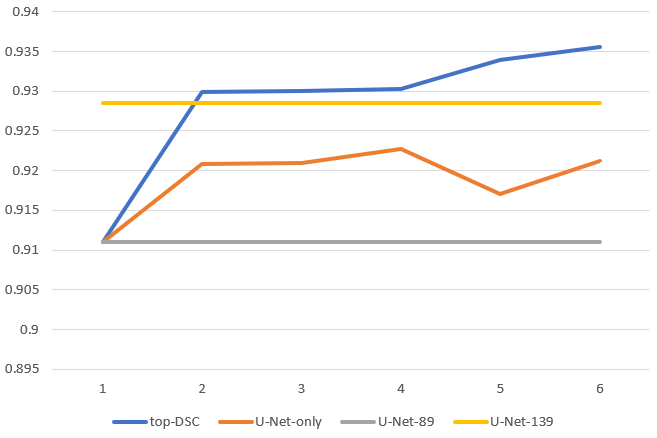}
    \caption{Iterative procedure of self-taught learning utilizing U-Net and Dense CRF for pseudo-labeling: (A) preprocessed input data; (B) masks manually labeled; (C) new data; (D) raw prediction; (E) refine prediction}
    \label{fig:result}
\end{figure}

An Overview of the final results in test data is shown in Table ~\ref{table_results} and Fig~\ref{fig:result}. The first column shows the results of selecting top 10 Dice's Coefficient masks in each iteration. And for the second column, in which, we just implemented U-Net without post-processing. Last two columns stand for the best U-Net performance with 89 and 139 ground truth masks respectively. Results show that top-DSC has the best performance with 89 ground truth masks and 50 pseudo labels. More interestingly, it even outperforms the results of U-Net trained with all the  139 well-annotated images. We hypothesise that the following two mechanisms contribute this phenomenon:1) the ``curriculum  Learning" strategy reduces the risk of local optimal, and 2) pseudo-labels provided by dense CRF insert certain level of noises into the learning process. This has been reported to be effective in reducing the risk of overfitting~\cite{ren2016ensemble}. In addition, we also observe that the proposed method still shows an uptrend by the end, while U-Net-only modeling is almost vibrating.
\begin{table}[htb]
\caption{RESULTS OF EXPERIMENTS}
\label{table_results}
\begin{center}

\begin{tabular}{lllll}
\hline
Iteration & top-DSC & U-Net-only & U-Net-89 & U-Net-139 \\ \hline
1         & 0.911   & 0.911     & 0.911   & -        \\
2         & 0.9299  & 0.92088   & -       & -        \\
3         & 0.93008 & 0.92093   & -       & -        \\
4         & 0.93031 & 0.92279   & -       & -        \\
5         & 0.93393 & 0.91709   & -       & -        \\
6         & 0.93562 & 0.92123   & -       & 0.92853  \\ \hline
\end{tabular}
\end{center}
\end{table}

\section{CONCLUSIONS} \label{conclusions}
 
 This paper explores a novel semi-supervised self-taught deep learning method for finger bones segmentation.  The proposed method utilizes a deep U-Net as the student module and a dense CRF as a teacher module. The student module is first initialized on a limited number of training set. Then the system is able to boost itself in a ``Curriculum  Learning"  manner. More specifically, in each iteration, the CRF refines the raw predictions of U-Net and return the results from the easiest samples therein as pseudo-labels for future learning. Experimental results show that the proposed method outperforms conventional supervised learning approaches.

\addtolength{\textheight}{-12cm}

\section*{ACKNOWLEDGMENT}
The work was supported by Singapore-China NRF-NSFC Grant [No. NRF2016NRF-NSFC001-111].

\bibliographystyle{IEEEbib}
\bibliography{refs.bib}

\end{document}